\pdfoutput=1

\documentclass[11pt]{article}

\usepackage[final]{acl}

\usepackage{times}
\usepackage{latexsym}
\usepackage{amsfonts}
\usepackage{arydshln}

\makeatletter
\def\eg{\textit{e.g.}} 
\def\ie{\textit{i.e.}}

\makeatother

\newcommand{\cut}[1]{}

\usepackage[T1]{fontenc}

\usepackage[utf8]{inputenc}

\usepackage{microtype}

\usepackage{inconsolata}

\usepackage{graphicx}

\title{On-device System of Compositional Multi-tasking\\ in Large Language Models}

\author{
 \textbf{Ondrej Bohdal\textsuperscript{1}},
 \textbf{Konstantinos Theodosiadis\textsuperscript{2}},
 \textbf{Asterios Mpatziakas\textsuperscript{2}},
 \\
 \textbf{Dimitris Filippidis\textsuperscript{2}},
 \textbf{Iro Spyrou\textsuperscript{2}},
 \textbf{Christos Zonios\textsuperscript{2}},
 \textbf{Anastasios Drosou\textsuperscript{2}},
 \\
 \textbf{Dimosthenis Ioannidis\textsuperscript{2}},
 \textbf{Kyeng-Hun Lee\textsuperscript{3}},
 \textbf{Jijoong Moon\textsuperscript{3}},
 \textbf{Hyeonmok Ko\textsuperscript{3}},
 \\
 \textbf{Mete Ozay\textsuperscript{1}},
 \textbf{Umberto Michieli\textsuperscript{1}}
\\
\\
 \textsuperscript{1}Samsung R\&D Institute UK, United Kingdom,
 \textsuperscript{2}CERTH, Greece,\\
 \textsuperscript{3}Samsung Research, South Korea
\\
 \small{
   \textbf{Correspondence:} \href{mailto:o.bohdal.1@samsung.com}{o.bohdal.1@samsung.com}
 }
}

\begin{document}
\maketitle

\newcommand{\um}[1]{\textcolor{blue}{UM: #1}}
\newcommand{\ob}[1]{\textcolor{green}{OB: #1}}

\begin{abstract}
Large language models (LLMs) are commonly adapted for diverse downstream tasks via parameter-efficient fine-tuning techniques such as Low-Rank Adapters (LoRA). While adapters can be combined to handle multiple tasks separately, standard approaches struggle when targeting the simultaneous execution of complex tasks, such as generating a translated summary from a long conversation. To address this challenge, we propose a novel approach tailored specifically for compositional multi-tasking scenarios involving summarization and translation. Our technique involves adding a learnable projection layer on top of the combined summarization and translation adapters. This design enables effective integration while maintaining efficiency through reduced computational overhead compared to alternative strategies requiring extensive retraining or sequential processing. We demonstrate the practical viability of our method within an on-device environment by developing an Android app capable of executing compositional tasks seamlessly. Experimental results indicate our solution performs well and is fast in both cloud-based and on-device implementations, highlighting the potential benefits of adopting our framework in real-world applications demanding high-speed operation alongside resource constraints.
\end{abstract}

\begin{figure}
  \includegraphics[trim=0cm 0cm 0cm 0cm,clip,width=\columnwidth]{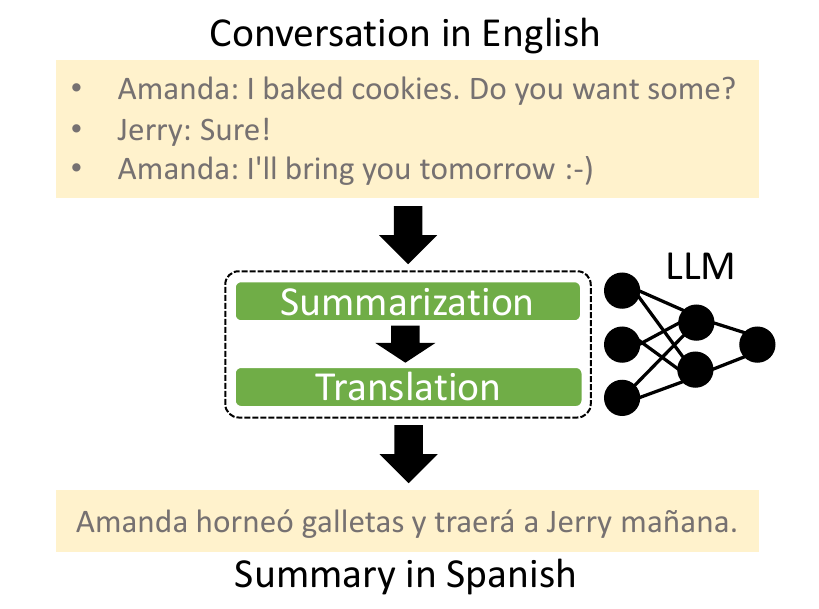}
  \caption{Compositional multi-tasking on the combination of summarization and translation. In our fully on-device system, we focus on the scenario where a conversation \cite{gliwa2019samsum} in one language is summarized in another language.}
  \label{fig:overview}
\end{figure}

\section{Introduction}

Generative AI has gained significant attention thanks to its ability to generate useful content \cite{gozalo2023survey} across modalities, including text \cite{zhao2023survey,minaee2024large}, images \cite{yang2024diffusion, cao2024diffusion, shenaj2024lora} and videos \cite{zhou2024survey}. 
Most generative AI applications to date rely on remote servers with advanced hardware. 
Nevertheless, there has been growing interest in harnessing on-device generative AI capabilities \cite{xu2024device}. On-device AI offers several advantages, particularly enhanced privacy since sensitive data remain securely stored on the device without transmission over networks \cite{dhar2021ondevice}. 
For text-based applications, compact yet proficient language models (LLMs)--typically ranging from 1B to 3B parameters--have emerged as viable options for deployment on mobile devices \cite{xu2024device}. Fine-tuning these pre-trained models using low-rank adaptation (LoRA) techniques  \cite{hu2021lora} significantly boosts their effectiveness across various tasks such as translation and summarization \cite{mao2025survey}.

A recent practical application of on-device LLMs concerns the so-called \textit{compositional multi-tasking}, which entails performing multiple tasks simultaneously \cite{bohdal2025compositional}.
Examples include generating translated summaries or adjusting tones in message replies.
To tackle this challenge, researchers proposed a learnable calibration mechanism that combines the corresponding LoRA parameters and then corrects the combination via a small number of additional parameters. 
Remarkably, these supplementary components require minimal storage space relative to standalone LoRAs (\eg, 0.5\%). We propose and implement an alternative efficient strategy that adds a projection layer on top of combined adapters.
However, our primary goal is not to propose a new method that surpasses state-of-the-art compositional multi-tasking approaches in \cite{bohdal2025compositional}.

In this paper, our main goal is to develop a fully on-device system implementing the compositional multi-tasking setup and to provide an associated on-device evaluation. We discuss details of how we developed an application running fully on-device for this conceptual setting, how the application works, and an experimental comparison between server and smartphone settings. More specifically, we focus on the combination of summarization and translation tasks, in an applied setting where we summarize messages that users receive on their phones. The considered scenario is illustrated in Figure~\ref{fig:overview}. We note that this is a highly practical use-case as it can be beneficial, for example, when people move abroad and join local chat groups that use the local language. The tool enables users to easily see the summary of the conversation in their own language.

\section{Related Work}
\subsection{On-device LLMs}
Large Language Models (LLMs) typically include billions of parameters, with the largest models currently containing over 400B+ parameters \cite{dubey2024llama}. Running these models, even only for inference, is costly and requires multiple high-end GPUs \cite{borzunov2024distributed}. In practice, this necessitates sending data to remote servers for generating suitable outputs. However, LLMs would also be highly beneficial in cases where private data (\eg,~messages) are used and when users prefer not to send data to the cloud \cite{dhar2021ondevice}. As a result, smaller LLMs of sizes such as 1B or 3B have been developed, making it possible to deploy LLMs directly on mobile devices. In these cases, all computations run locally, avoiding data transmission to remote servers and also reducing operational costs for service providers. Various models suitable for on-device deployment have been developed, for example, LLaMA 3.2 1B \cite{dubey2024llama}, Qwen2.5 1.5B \cite{qwen2,qwen2.5}, StableLM2 1.6B \cite{bellagente2024stable} and Gemma 2B \cite{team2024gemma}.

\subsection{Multi-tasking in LLMs}
LLMs can perform diverse tasks after their large-scale pre-training \cite{zhao2023survey,minaee2024large}, but for strong performance on individual tasks, they are typically fine-tuned via parameter-efficient fine-tuning (PEFT) \cite{han2024parameter,ding2022delta}. Fine-tuning is especially beneficial for on-device LLMs that have more limited resources. A common strategy for PEFT is the use of low-rank adapters (LoRA) that are injected into selected layers \cite{hu2021lora}, introducing only a comparatively small number of parameters (\eg,~10M for a model of size 1B). These LoRA parameters are then loaded into a shared LLM to perform individual tasks \cite{mao2025survey}. It is common to store the single-task LoRAs on the device alongside the base LLM model \cite{gunter2024apple}. In order to perform multi-tasking, the fine-tuned models can be merged into each other. Various strategies exist, including simple linear merging that computes a weighted average of the weights \cite{wortsman2022model,ilharco2022editing}, TIES merging \cite{yadav2024ties}, and more advanced learnable strategies such as LoraHub \cite{huang2023lorahub}. Such model merging strategies have been shown to work well when doing multiple tasks separately \cite{yang2024model}. However, they have been shown not to work well in compositional multi-tasking where specialized approaches need to be developed. Well-performing na\"ive baselines for compositional multi-tasking are inefficient as they either require training a new specialized LoRA or performing two inference passes with the LLM in sequence. In this work, we develop an on-device system that integrates compositional multi-tasking.

\section{Problem Statement and Method}
We focus on compositional multi-tasking, where two tasks are combined to be performed jointly. More specifically, we consider the summarization of messages in English as the primary task $T_1$ with the secondary task $T_2$ of translation from English to Spanish. The compositional task performs $T^C_{1,2}(x)=T_2(T_1(x))$, where $x \mapsto y_1 \mapsto y_2$ and a generic task $T$ takes input text $x$ and outputs text $y$; \ie, $T(\cdot): x \mapsto y$.

We assume that an LLM model and LoRAs for tasks $T_1, T_2$ parameterized by $B_1, A_1$ and $B_2, A_2$ are already stored on the device. More specifically, LoRA \cite{hu2021lora} introduces low-rank factorized matrices ${B \in \mathbb{R}^{d\times r}}$, $A \in \mathbb{R}^{r\times k}$ where the rank $r \ll \min(d, k) $. Here, parameters $k, d$ specify the input and output dimensions for the given layer respectively.
Then, $A$ and $B$ are combined with model's weights $W_0 \in \mathbb{R}^{d\times k}$, resulting in an adjusted forward pass:
\begin{equation}
    h=(W_0 +\Delta W) x=(W_0+BA)x.
\end{equation}

Various baselines can be considered:
(i) a \textit{zero-shot} approach where the model is prompted about the task, 
(ii) either the \textit{primary-task LoRA} or the \textit{secondary-task LoRA} paired with prompting, 
(iii) various merging strategies applied on the LoRA parameters (\eg,~\textit{linear} \cite{wortsman2022model}, \textit{concatenation} \cite{peft}, \textit{TIES} \cite{yadav2024ties} and \textit{LoraHub} \cite{huang2023lorahub},
(iv) inefficient baselines that perform well: \textit{two-step LoRA usage} and \textit{joint-expert LoRA} for the specific compositional task.

The goal is to obtain performance comparable to the inefficient baselines while being more efficient, in particular only requiring one inference pass and introducing only a limited number of additional parameters instead of a new LoRA that would use \eg~50MB in storage.

\begin{figure}[t]
  \centering
  \includegraphics[width=\linewidth]{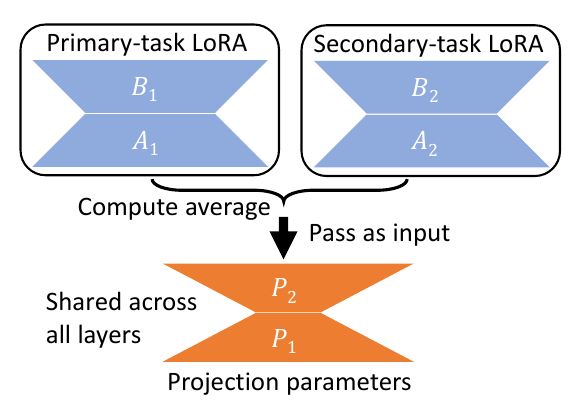}
  \caption{Overview of our projection merge method. We first merge the single-task LoRAs and then pass them through projection parameters.}
  \label{fig:proj_merge}
\end{figure}

We propose a new strategy that learns very few additional specialized parameters, using data $\mathbb{D}^C$ from the compositional task $T^C_{1,2}$. The inputs $x$ are conversations in English, while the targets $y$ are ground-truth summaries that have been translated from English to Spanish via a specialized translation model. These parameters are pre-trained on a server and then deployed to the device.
Our technique adjusts the forward pass $h=(W_0 +\Delta W) x$ with update matrix $\Delta W$ computed as:
\begin{equation}
\Delta W = P_2 P_1 (0.5 B_1 A_1 + 0.5 B_2 A_2).    
\end{equation}
We refer to it as \textit{projection merge} because it projects the average of the individual LoRA parameters via additional low-rank parameters $P_2 \in \mathbb{R}^{d\times s}, P_1 \in \mathbb{R}^{s\times k}$ for $s \ll \min(d, k)$. These projection parameters are shared across layers as well as components that have the same input and output dimensions (\eg~key, value attention projections). An overview of the method is provided in Figure~\ref{fig:proj_merge}.

Our solution has negligible storage overhead, unlike the \textit{joint-expert LoRA} approach that requires storing a new LoRA. Further, the compute requirements are similar for both as our approach involves multiplying one LoRA matrix by the projection matrix. This introduces only negligible additional cost, as the overall runtime is dominated by inference with the base model. 

\section{System Framework}
\label{sec:System}

\subsection{Proof-of-Concept Application on Server}
We developed our application in two phases. First, we created a proof-of-concept (PoC) application running on a server using Python and React Native. The system consists of a React Native user client and a FastAPI server that manages communications between the user and the LLM. We used the PEFT Python library \cite{peft} to load a Llama-3.2-1B-Instruct \cite{dubey2024llama} model and the LoRA adapters, while the Transformers library \cite{wolf2019huggingface} handled the tokenizer and end-to-end LLM pipeline.

\begin{figure}[ht]
  \centering
  \includegraphics[width=0.493\linewidth]{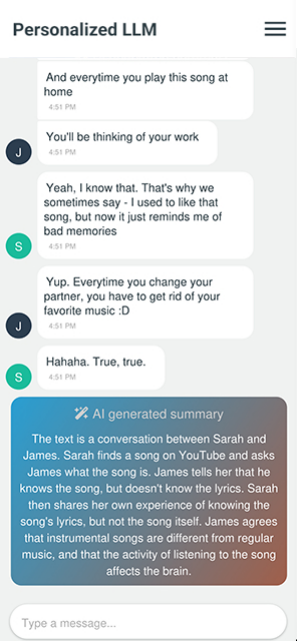}
  \includegraphics[width=0.493\linewidth]{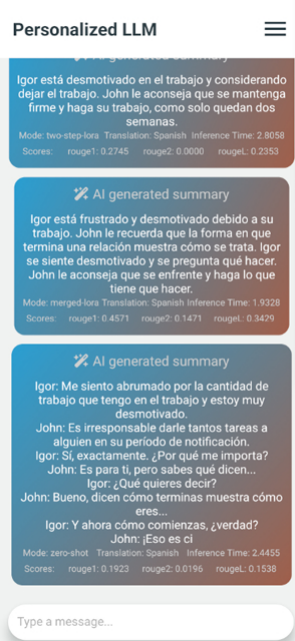}
  \caption{The user client for the PoC server-side application. The application operates either in a predefined mode such as two-step LoRA usage (left) or in an experimental mode where all methods are utilized and relevant metrics are reported (right). Tapping on the summary translates it back to English, which has been used in the example on the left side.
  }
  \label{fig:python_app}
\end{figure}

The application supports three methods for the joint task of summarization and translation: (i) our proposed projection merging method (\textit{Merged LoRA}); (ii) base LLM model with prompting (\textit{zero-shot}); and (iii) sequential LoRA adapter use (\textit{two-step}) where one adapter is loaded to handle summarization and a second one later to handle translation of the summary. The system can also run automated experiments comparing all three methods, calculating ROUGE scores \cite{lin2004rouge} and inference times. Figure~\ref{fig:python_app} demonstrates the two-step method (left) and a comparison of all three methods (right). For easier English-reader assessment, the summary in the left panel has been translated back to English. All example conversations are from the SAMSum dataset \cite{gliwa2019samsum}.

\subsection{On-device Application}
\subsubsection{Framework Components}

\begin{figure}[h]
  \centering
  \includegraphics[width=\linewidth]{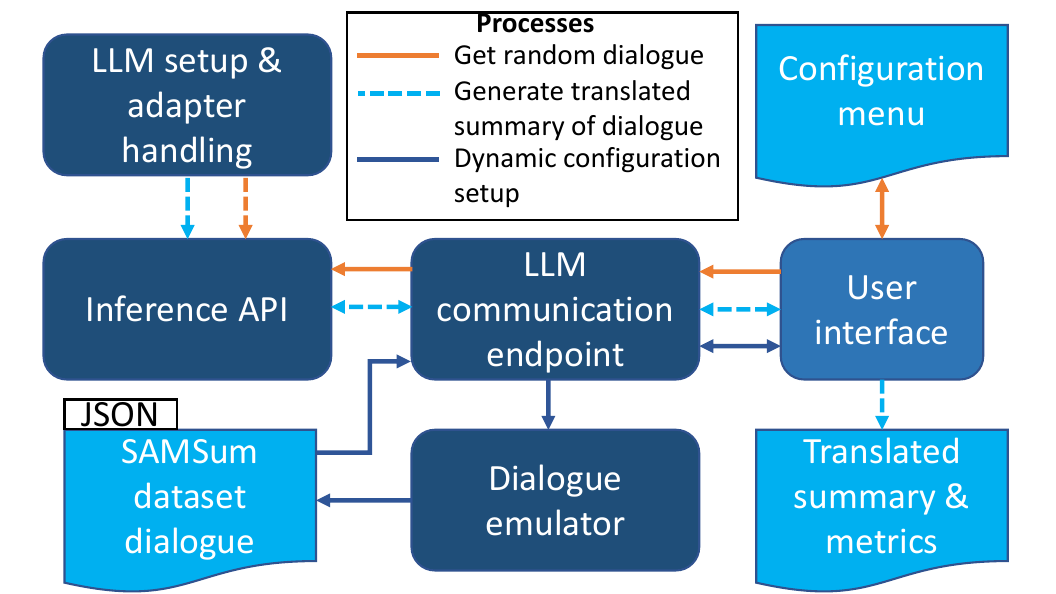}
  \caption{High-level architecture of our on-device system for compositional multi-tasking.}
  \label{fig:rust_app_architecture}
\end{figure}

For the on-device implementation, we built the back-end in Rust and the front-end in React Native. We used the \textit{mistral.rs} package \cite{mistral_rs} for LLM and adapter management, and the \textit{axum crate} \cite{axum_crate} for front-end/back-end communication. 
Figure~\ref{fig:rust_app_architecture} shows the system's high-level architecture. We describe each component of the architecture next.
The \textit{user interface} either receives an input dialogue from the user or can load example dialogues from the dataset so that the user does not need to enter long conversations. 
It communicates with the \textit{LLM communication end-point}. The functionalities offered by the user interface are elaborated in Subsection \ref{ssec:UI}. 
The LLM communication endpoint handles all communications between the front-end and back-end of the application, \eg~it is connected to the \textit{dialogue emulator} that loads example dialogues and realistically replays them. 
It is also connected to the \textit{inference API} that is responsible for handling the LLM prompt, calculation of metrics, and most importantly doing the actual inference that outputs the translated summary of the dialogue. Finally, \textit{LLM setup and adapter handling} tackles the task of loading the LLM model and adapters while allowing for the dynamic configuration of these components. 

\subsubsection{User Interface}
\label{ssec:UI}
\begin{figure}[h]
  \centering
  \includegraphics[width=0.493\linewidth]{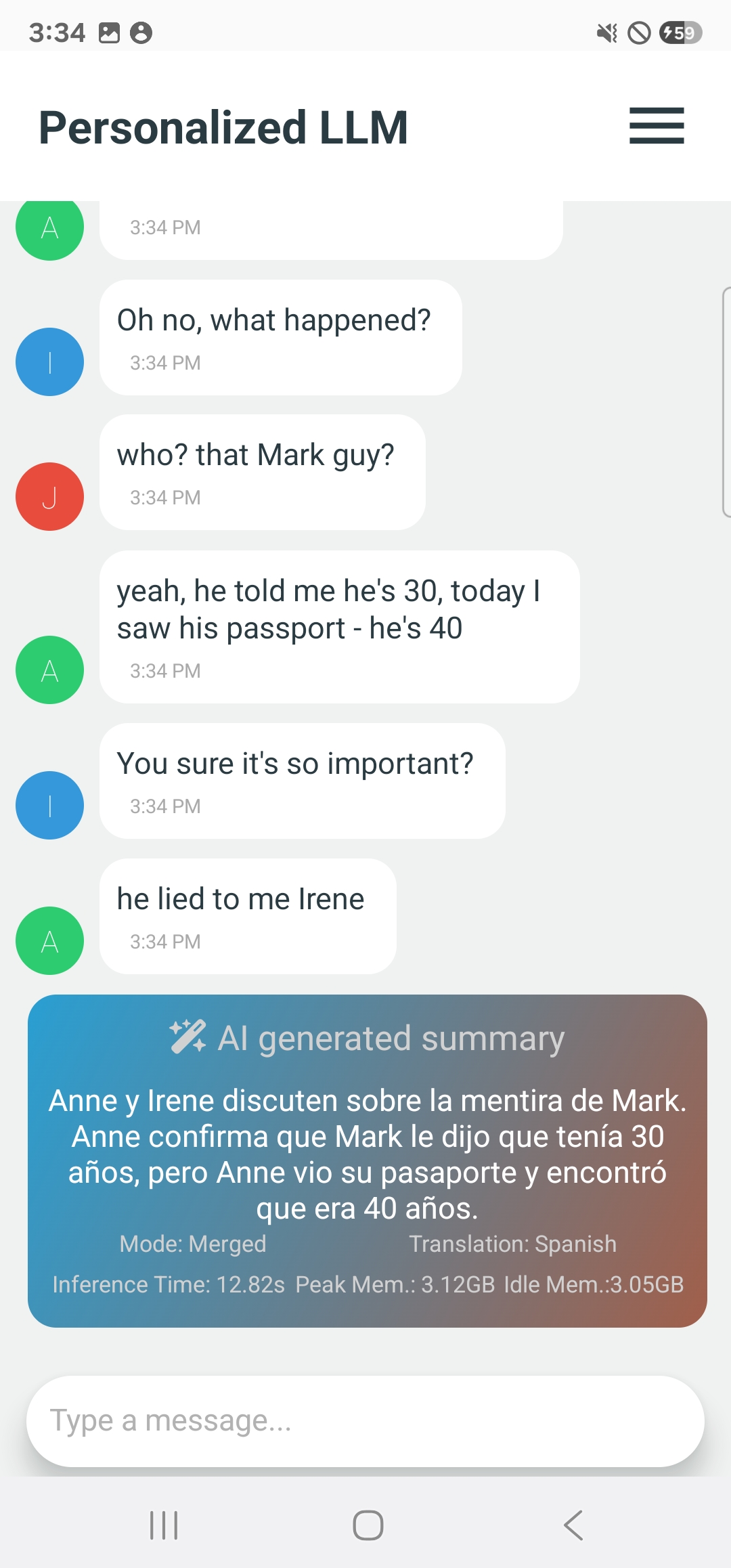}
  \includegraphics[width=0.493\linewidth]{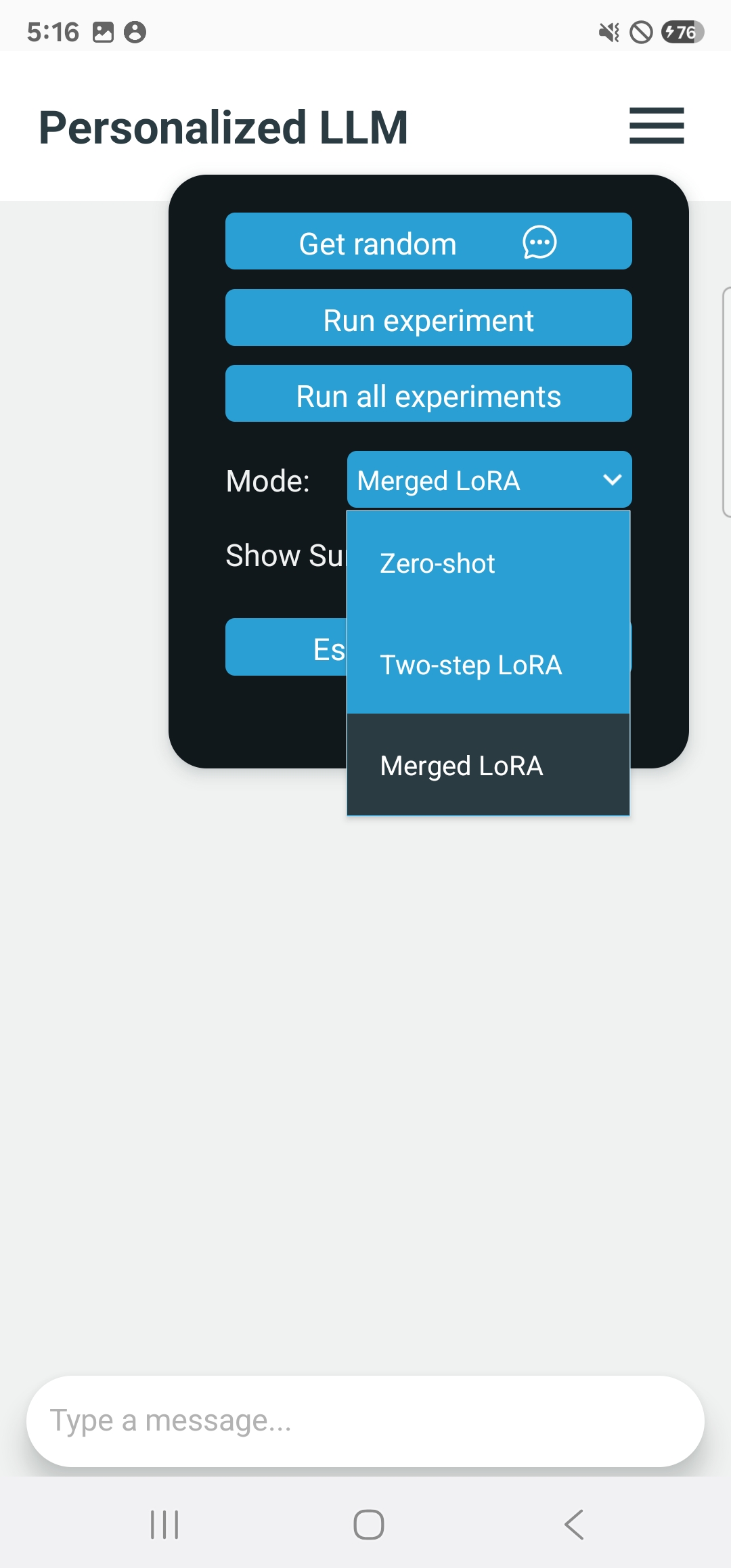}
  \caption{The application runs entirely on a Samsung S23 Ultra Android device. While similar to the server-side proof-of-concept application, additional information is displayed such as memory consumption. }
  \label{fig:rust_app}
\end{figure}

The user interface provides a configuration menu and displays the translated summary and metrics below the conversation, as shown in Figure~\ref{fig:rust_app}. Key features include:
(i) method selection;
(ii) task configuration, \ie, whether to run summarization only or with translation;
(iii) optional random conversation generation;
(iv) single or comparative method evaluation mode, \ie, whether to run one experiment with the selected method or with all methods;
(v) target language selection.

\subsubsection{Implementation Challenges and Solutions}

Compared to a PoC solution on the server, developing an on-device LLM system presented several technical challenges, including the following:

1) Adapter Integration: to handle this challenge, we modified the \textit{mistral.rs} inference library. We exported the relevant classes from the library, and modified them to directly use local adapters. Consequently we were able to load the adapters within our Android app.

2) Model Loading: the \textit{mistral.rs} library supports the \textit{GGUF} quantized model format \cite{ggml} with LoRA weights for Llama models. However, by default, it merges the LoRA weights into the base weights of the model upon loading. This behavior was undesirable for two reasons: if the weights are integrated into the model directly, we do not have the ability to dynamically switch between several adapters; further, in order to facilitate the suggested approach, the application should apply a custom merging of LoRA weights only. To address this, we modified the library's loading mechanism to bypass the default merging of LoRA weights, allowing greater control over adapter management.

3) Memory Management: another significant issue was related to using the main user interface (UI) thread to launch the inference back-end. Specifically, the application would not load the inference engine due to its high memory requirements. To overcome this hurdle, we moved all processing tasks to a separate thread dedicated to I/O operations using Kotlin, ensuring heavy tasks would not block the UI thread. This allowed the UI thread to remain free for loading the UI without delays and for handling user interactions. This approach improved the overall performance and responsiveness of the application and, more importantly, allowed the inference engine to function correctly.

\section{Experiments}
\label{sec:Experiments}

\subsection{Implementation Details}

We use Llama-3.2-1B-Instruct model \cite{dubey2024llama} with individual LoRA adapters trained on the SAMSum conversation summarization dataset \cite{gliwa2019samsum} and TEDTalks English-to-Spanish translation dataset \cite{qi2018pretrained}. For the on-device system, Q4\_K\_M 4-bit quantization of the model was used \cite{llama_3_2_1b_instruct_gguf}. For the combined summarization-translation task, we created ground-truth data by translating SAMSum summaries using the Opus Machine Translation model \cite{tiedemann2023democratizing,TiedemannThottingal2020opus}. The statistics for each task are specified in Table~\ref{tab:dataset_statistics}.
We chose the SAMSum dataset because it aligns well with our target use case of cross-lingual summarization on mobile devices. SAMSum features written-style conversations, which reflects well the input expected in our application. In contrast, the DialogSum dataset \cite{chen2021dialogsum} used in \cite{bohdal2025compositional} focuses on spoken-style dialogues. 

\begin{table}[h!]
\caption{Dataset statistics across the different tasks.}
\label{tab:dataset_statistics}
\begin{center}
\resizebox{\columnwidth}{!}{
\begin{tabular}{lccc}
\hline
& \textbf{Training} & \textbf{Validation} & \textbf{Test} \\
\hline
Summarization & \phantom{0}14,732 & \phantom{0,}818 & \phantom{0,}819 \\
Translation & 196,026 & 4,231 & 5,571 \\
Cross-lingual summarization	 & \phantom{0}14,732 & \phantom{0,}818 & \phantom{0,}819 \\
\hline
\end{tabular}}
\end{center}
\end{table}

LoRAs are applied to attention components (query, key, value, output projections) and multi-layer perceptron (MLP) components (up, down, gate projections) \cite{fomenko2024note,Tunstall2024recipes}. We train them using the Adam optimizer with a learning rate of $5\times 10^{-5}$ and minibatch size of 3 for one epoch on the full training set. The LoRAs use rank $r=32$, parameter $\alpha=16$, dropout rate 0.05, resulting in 22.5M parameters and 45.1MB storage per adapter.

The parameters of our projection merge approach are trained using the Adam optimizer with learning rate $5\times 10^{-4}$ and minibatch size of 3, training for one epoch on 10,000 randomly selected examples from the training dataset. The examples used for training are conversations in English, while the targets are the ground-truth summaries translated from English to Spanish. We use rank $s=4$, resulting in 0.1M additional parameters (0.2MB storage). For the alternative merging strategies, we selected weights of 0.5 for Linear, Concat, and TIES merges, and a density of 0.5 for the TIES merge. For all compared approaches, we include a system prompt specifying the task: \textit{``Summarize the following text and translate it from English to Spanish''}.

\subsection{Performance Analysis}
As an initial step, we have performed experiments on a server with GPUs to compare our projection merge approach against the different baselines. The results are reported in Table~\ref{tab:server} and we can extract the following key findings.
(i) A simple zero-shot strategy obtains poor performance despite including a prompt that specifies the compositional multi-task objective. 
(ii) Primary or secondary-task LoRAs as well as various merging strategies perform better on the compositional task examined, but their performance is significantly lower than that of the two-step LoRA usage or joint-expert LoRA.
(iii) Our proposed projection merge achieves better performance than both two-step LoRA usage and joint-expert LoRA baselines.
The efficiency of our projection merge is analyzed in Table~\ref{tab:efficiency}, showing that our solution introduces only 0.4\%  of the parameters compared to a specialized joint-expert LoRA while requiring just one inference step.

\begin{table}[h!]
\caption{Evaluation on summarization of conversations in another language, test ROUGE-1/2/L (\%, $\uparrow$). Our projection merge obtains comparable (slightly better) performance to the inefficient two-step LoRA usage or joint-expert LoRA that trains a full adapter for the given compositional task. Other baselines obtain significantly weaker performance.}
\label{tab:server}
\begin{center}
\resizebox{\columnwidth}{!}{
\begin{tabular}{lccc}
\hline
& \textbf{ROUGE-1} & \textbf{ROUGE-2} & \textbf{ROUGE-L} \\
\hline
Zero-shot & 18.55 & \phantom{0}4.60 & 13.72 \\
Primary-task LoRA & 23.14 & \phantom{0}6.37 & 17.62 \\
Secondary-task LoRA & 28.08 & \phantom{0}8.08 & 20.94 \\
\hdashline
Linear merge & 27.38 & \phantom{0}7.73 & 20.34 \\
Concat merge & 27.58 & \phantom{0}7.84 & 20.45 \\
TIES merge & 25.16 & \phantom{0}6.91 & 18.36 \\
LoraHub merge & 28.05 & \phantom{0}7.90 & 20.82 \\
\hdashline
Two-step LoRA usage & 37.26 & 13.64 & 29.25 \\
Joint-expert LoRA & 35.10 & 11.56 & 26.99 \\
\hdashline
Projection merge (ours) & 37.21 & 14.41 & 30.21 \\
\hline
\end{tabular}}
\end{center}
\end{table}

\begin{table}[h!]
\caption{Efficiency of our solution vs inefficient but well-performing baselines. Our method needs only one inference pass and just 0.4\% of the parameters and storage compared to a new LoRA.}
\label{tab:efficiency}
\setlength{\tabcolsep}{3.5pt}
\begin{center}
\resizebox{\columnwidth}{!}{
\begin{tabular}{lccc}
\hline
\textbf{Method} & \multicolumn{1}{p{2cm}}{\centering \textbf{Number of}  \\ \textbf{Inferences}} &  \multicolumn{1}{p{2cm}}{\centering \textbf{Additional} \\ \textbf{Parameters}} & \multicolumn{1}{p{2cm}}{\centering \textbf{Additional} \\ \textbf{Storage}} \\
\hline
Two-step LoRA usage & $2\times$ & \phantom{0}0.0M & \phantom{0}0.0MB \\
Joint-expert LoRA & $1\times$ & 22.5M & 45.1MB \\
\hdashline
Projection merge (ours) & $1\times$ & \phantom{0}0.1M & \phantom{0}0.2MB \\
\hline
\end{tabular}
}
\end{center}
\end{table}

We note that the primary goal of our work is to detail how to develop an on-device system for compositional multi-tasking, without aiming to outperform state-of-the-art solutions for compositional multi-tasking. 
For a more complete context, we include a comparison with Learnable Calibration \cite{bohdal2025compositional} on the task utilized in our paper. Results in Table~\ref{tab:learnable_calibration_comparison} and \ref{tab:learnable_calibration_efficiency} show that our projection merge achieves a balance between performance and efficiency, lying between the two Learnable Calibration variants.

\begin{table}[h!]
\caption{Evaluation on summarization of conversations in another language, test ROUGE-1/2/L (\%, $\uparrow$). The performance of our projection merge lies between the two Learnable Calibration variants.}
\label{tab:learnable_calibration_comparison}
\begin{center}
\resizebox{\columnwidth}{!}{
\begin{tabular}{lccc}
\hline
& \textbf{ROUGE-1} & \textbf{ROUGE-2} & \textbf{ROUGE-L} \\
\hline
Learnable Calibration & 32.87 & 12.53 & 26.54 \\
Learnable Calibration++ & 39.84 & 16.66 & 32.63 \\
\hdashline
Projection merge (ours) & 37.21 & 14.41 & 30.21 \\
\hline
\end{tabular}}
\end{center}
\end{table}

\begin{table}[h!]
\caption{Efficiency of our solution vs Learnable Calibration. The efficiency of our projection merge lies between the two Learnable Calibration variants.}
\label{tab:learnable_calibration_efficiency}
\setlength{\tabcolsep}{3.5pt}
\begin{center}
\resizebox{\columnwidth}{!}{
\begin{tabular}{lccc}
\hline
\textbf{Method} & \multicolumn{1}{p{2cm}}{\centering \textbf{Number of}  \\ \textbf{Inferences}} &  \multicolumn{1}{p{2cm}}{\centering \textbf{Additional} \\ \textbf{Parameters}} & \multicolumn{1}{p{2cm}}{\centering \textbf{Additional} \\ \textbf{Storage}} \\
\hline
Learnable Calibration & $1\times$ & \phantom{0}24K & 0.05MB \\
Learnable Calibration++ & $1\times$ & 176K & 0.35MB \\
\hdashline
Projection merge (ours) & $1\times$ & 102K & 0.20MB \\
\hline
\end{tabular}
}
\end{center}
\end{table}

\subsection{System Analysis}

\begin{figure}[h]
  \centering
  \includegraphics[width=\columnwidth]{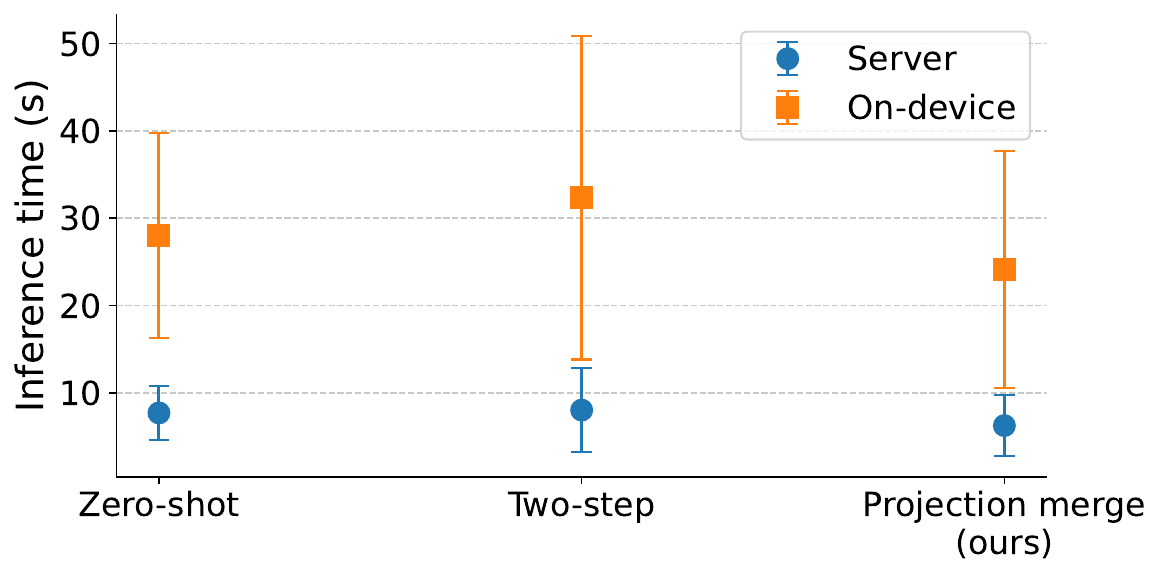}
  \caption{Comparison of inference times for selected approaches on the server and on the device, with mean and standard deviation across the trials. Our projection merge is faster than the baselines, taking around 6 seconds on the server and 24 seconds on the device.}
  \label{fig:time}
\end{figure}

We have performed experiments on a subset of data (using about 20\% of the test conversations to make the on-device evaluation faster) to compare inference times on the server and on the device. Our on-device experiments utilize a Samsung S23 Ultra Android device, while our server-side experiments in this section utilize an AMD Ryzen 7, 16 cores CPU. We compare \textit{zero-shot}, \textit{two-step LoRA}, and our \textit{projection merge} approach in the evaluation. 
Figure~\ref{fig:time} shows our method achieves the fastest inference times, taking around 6 seconds on the server and 24 seconds when running on the device. The standard deviations are relatively larger as the inference time depends on the length of the conversation and of the resulting summary.

While GPU inference reduces these times to under 1 second, our experimental evaluation suggests that the time required when performing all computations on the device is manageable, even though it could be optimized further by exploring more advanced quantization methods
such as BitNet \cite{wang2023bitnetscaling1bittransformers}.
The observed speedup of our method is comparatively larger when running on-device in contrast to when it runs on the server.

We remark the time employed by the two-step solution is not twice as long as of the other approaches because the inference time depends on the number of processed and generated tokens. The first inference pass performs summarization and so the input for the second inference pass is significantly shorter. When an approach generates long outputs, it takes a proportionally longer time, so the overall inference time depends strongly on how much the solution leads to generating long outputs.

Analysis of memory has shown all methods require around 3.12GB of peak memory, with 3.01GB being the mean idle memory requirement after loading the model and adapters. Hence most memory is used for loading the model and adapters, while using the model for inference consumes only a small amount of additional memory.

\subsection{Discussion}
Our work establishes the feasibility of running compositional multi-tasking LLMs entirely on-device, ensuring user privacy by eliminating the need for remote server communication. The modular design of our application supports straightforward extension to additional languages and compositional tasks. For example, additional compositional tasks could include reply suggestions in addition to summarization and their combinations with translation and tone adjustment. 

The current implementation faces some practical limitations, namely: inference times of over 20 seconds may be too long for some use cases; memory requirements of ~3GB may make deployment challenging for tiny remote devices.
These limitations could be addressed through more aggressive quantization techniques, model architectures optimized for mobile devices, shared parameters across adapters, and dynamic adapter loading based on user needs. However, the most promising avenue for significant speedups would be running an LLM integrated into the mobile operating system, rather than within our application.

\section{Conclusion}

This paper presented an on-device system for compositional multi-tasking in LLMs, focusing on the practical use case of summarizing conversations in another language. This capability is particularly valuable for users engaging with foreign language content, such as travelers participating in local chat groups. Our solution introduces a lightweight projection layer on top of single-task LoRAs, achieving superior performance with minimal parameter overhead compared to efficient but poorly-performing baselines or well-performing but inefficient approaches.
Experimental evaluation on a smartphone has confirmed the practical benefits of our solution, highlighting its speed and viability in fully on-device settings.

\section*{Limitations}
While we show it is possible to use our method in fully on-device settings, inference times of over 20 seconds may be considered too long. As a result, further optimization of the system, \eg~via more aggressive quantization, would be needed before wider deployment. The system requires a manageable amount of memory (3GB), but even this could still make it suitable only for mid and higher-end devices. The solution requires additional parameters that are specific to the given compositional task, so for each new compositional task we would store these on the device. However, the amount of additional storage is small and negligible compared to storing a full adapter, making it possible to scale also to larger numbers of compositional tasks.

\section*{Ethical Considerations}
Ability to perform compositional multi-tasking fully on-device has significant practical benefits for users. User's data remain fully private and no connection to the internet is needed. While this has broad benefits for users, it also has implications for cases when the users would want to use the compositional multi-tasking abilities to achieve undesirable goals. While the original LLM model may have been aligned for safety, fine-tuning it via LoRA and performing subsequent LoRA merging can diminish the robustness of the safeguard mechanisms.

\bibliography{references}

\end{document}